\documentclass[10pt,twocolumn,letterpaper]{article}
\usepackage[pagenumbers]{cvpr} 
\pdfoutput=1

\usepackage[dvipsnames]{xcolor}

\def\tableimagewidth{0.22\columnwidth}
\def\tableimageheight{0.20\columnwidth}

\def\heidarihybridtableimagewidth{0.21\columnwidth}
\def\heidarihybridtableimageheight{0.20\columnwidth}

\def\ctoftableimagewidth{0.40\columnwidth}
\def\ctoftableimageheight{0.35\columnwidth}

\def\hteasertableimagewidth{0.312\columnwidth}
\def\teasertableimagewidth{0.317\columnwidth}

\newcommand{\NPhard}{\text{$\mathcal{NP}$-hard}}

\usepackage{algpseudocode}
\usepackage{algorithm}
\usepackage{braket}

\usepackage{amsmath, nccmath}
\usepackage{bm}
\usepackage{array,multirow,graphicx}
\usepackage{adjustbox}

\usepackage{widetext}
\usepackage{minitoc}

\usepackage{xcolor} 
\usepackage{svg}
\usepackage{adjustbox}
\definecolor{cvprblue}{rgb}{0.21,0.49,0.74}
\usepackage[pagebackref,breaklinks,colorlinks,citecolor=cvprblue]{hyperref}
\def\paperID{180}
\def\confName{3DV\xspace}
\def\confYear{2024\xspace}
\title{Quantum-Hybrid Stereo Matching\\With Nonlinear Regularization and Spatial Pyramids} 
\author{
Cameron Braunstein$^{1,2}$\hspace{2.3em}
Eddy Ilg$^{1}$\hspace{2.3em}
Vladislav Golyanik$^{2}$\vspace{0.3em} \\
$^1$Saarland University, SIC\vspace{0.3em} \hspace{1.3em}
$^2$MPI for Informatics, SIC
}
\begin{document}
\maketitle
\setcounter{page}{1}
\begin{abstract} 
Quantum visual computing is advancing rapidly. This paper presents a new formulation for stereo matching with nonlinear regularizers and spatial pyramids on quantum annealers as a maximum a posteriori inference problem that minimizes the energy of a Markov Random Field. Our approach is hybrid (\textit{i.e.,} quantum-classical) and is compatible with modern D-Wave quantum annealers, \textit{i.e.,} it includes a quadratic unconstrained binary optimization (QUBO) objective. Previous quantum annealing techniques for stereo matching are limited to using linear regularizers, and thus, they do not exploit the fundamental advantages of the quantum computing paradigm in solving combinatorial optimization problems. In contrast, our method utilizes the full potential of quantum annealing for stereo matching, as nonlinear regularizers create optimization problems which are \NPhard{}. On the Middlebury benchmark, we achieve an improved root mean squared accuracy over the previous state of the art in quantum stereo matching of 2\% and 22.5\% when using different solvers. 
\end{abstract} 
\begin{figure}
    \centering
     \setlength\tabcolsep{1.5pt}
     \renewcommand{\arraystretch}{0.7}
    \begin{tabular}{c c c}
        \includegraphics[width=\hteasertableimagewidth]{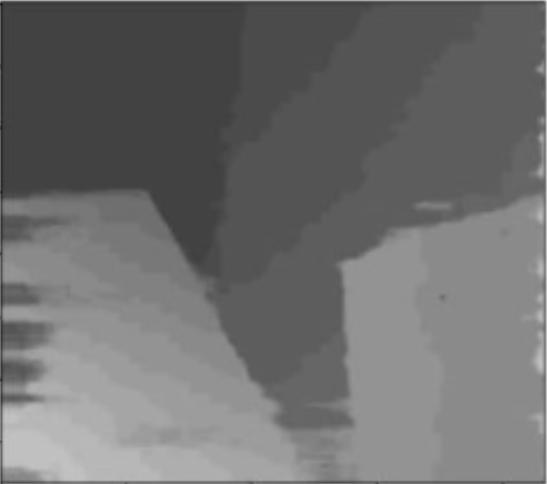} & \includegraphics[width=\teasertableimagewidth]{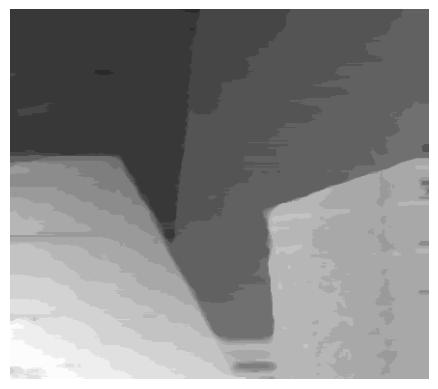}  & \includegraphics[width=\teasertableimagewidth]{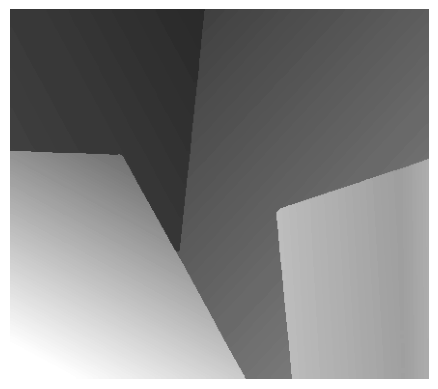} \\
        \begin{scriptsize}Prior Work  \end{scriptsize} &  \begin{scriptsize}Ours  \end{scriptsize}  &  \begin{scriptsize}Ground Truth \end{scriptsize} \\
         \begin{scriptsize} $\textit{RMSE}  = 1.4$  \end{scriptsize} & \begin{scriptsize} $\textit{RMSE} = 0.96$ \end{scriptsize} & \\
         \begin{scriptsize} $\textit{BPP} = 9.8$ \end{scriptsize} &  \begin{scriptsize} $\textit{BPP} = 8.16$ \end{scriptsize}& \\
    \end{tabular}
    \caption{Stereo estimates on the Middlebury \cite{middlebury2001} Venus stereo image pair. From left to right: Heidari \etal's approach \cite{9653310}, our approach and ground truth. In this example, we achieve a $46\%$ decrease in root mean squared error (\textit{RMSE}) from Heidari \etal.~and a $10\%$ decrease in bad pixel percentage (\textit{BPP}). We 
    avoid many of the streaking artifacts present in the result of the prior approach.} 
    \label{fig:teaser}
\end{figure} 
 
\section{Introduction}\label{sec:intro} 

Stereo matching has already been studied for more than a century \cite{Hauck1883, Finsterwalder1897, LonguetHiggins1981ACA}. It is well understood, and many algorithms exist \cite{disparity_graph_cuts,segment_based_stereo_matching,hierarchical_belief_propagation,hirschmuller2005accurate}, including recent works leveraging quantum hardware \cite{CruzSantos2018AQF, 9653310}. Despite the fact that quantum computers still cannot compete with classical machines in absolute terms (\textit{e.g.,} absolute execution speed or admissible problem sizes), they promise to provide accelerated solutions in certain cases, including combinatorial optimization problems in the near future. Quantum computing and quantum computer vision are quickly developing and gaining momentum \cite{Biondi2021, bravyi2022future, Larasati2022}. Hence, the community is investigating how such fundamental computer vision problems as stereo matching could benefit from quantum hardware.

The current leading method for quantum stereo matching from Heidari \etal ~\cite{9653310} is based on solving the min-flow-max-cut problem on a quantum annealer. In particular, quantum annealers are theorized to provide advantages in solving \textit{Quadratic Unconstrained Binary Optimization} (QUBO) problems and, \textit{e.g.,} outperform simulated annealing \cite{Kirkpatrick1983} for certain (rugged) energy landscapes, in terms of the absolute convergence speed and the energy level of the final returned solution \cite{PhysRevX.8.031016, YanSinitsyn2022}. While Heidari \etal ~\cite{9653310} achieve useful results, their method is limited to linear regularizers. Moreover, the solution of their approach can be computed in polynomial time and does not leverage the true advantages of quantum computers. Furthermore, it cannot process a complete epipolar line at a time on modern quantum hardware, and has no ability to process data with a large number of disparity labels via coarse-to-fine techniques. 

To address these shortcomings, we propose a novel stereo matching formulation for quantum annealers based on Markov Random Fields (MRFs). It allows modeling nonlinear regularizers that lead to $\mathcal{NP}$-hard problems and therefore exploits the true advantages of the quantum computing paradigm. In addition, it allows leveraging a coarse-to-fine pyramid for 
robust processing, see Fig.~\ref{fig:teaser}. A direct comparison to Heidari \etal~\cite{9653310} is challenging, as they use the D-Wave proprietary hybrid solver, and it is unknown to what degree their solution was obtained with traditional solvers or quantum hardware. When reimplementing their method and solving the max-flow-min-cut problem using Ford Fulkerson \cite{10.5555/1942094}, and comparing it to our method solved with the non-quantum solver Gurobi (\textit{i.e.,}~with a non-quantum branch-and-bound optimization), we achieve an RMSE improvement of $2\%$ on average. When directly comparing our Gurobi results to the results of their paper, the improvement is $22.5\%$. Thus, our method overall yields an improvement between $2\%$ and $22.5\%$. 

Even though solving the proposed objective on quantum hardware with D-Wave does not result in improved solutions yet due to imperfections of quantum hardware, the proposal of a method with nonlinear regularizers mappable to quantum hardware is a step forward. In summary, our contributions are as follows: 
\begin{itemize} 
\item We present a novel quantum-hybrid approach to stereo matching with MRF energy minimization that is compatible with modern and upcoming quantum hardware. 
\item For the first time, we show how nonlinear regularizers can be modeled for stereo matching on quantum annealers and the advantages of quantum computers on {\NPhard} problems can effectively be leveraged. 
\item We demonstrate that our approach allows integrating a coarse-to-fine pyramid that adds additional regularization and makes the computation tractable with current solvers. 
\end{itemize}
The source code of our approach is available, see our project website\footnote{\href{https://4dqv.mpi-inf.mpg.de/QHSM/}{https://4dqv.mpi-inf.mpg.de/QHSM/}} for details. 
The general MRF formulation for quantum annealer optimization can also be applied to other fundamental computer vision tasks, such as optical flow estimation or motion segmentation. 

\section{Related Work}\label{related_work}  

\subsection{Traditional Methods for Stereo Matching} 

Stereo matching is a fundamental task 
with a long history.
Most recent research focuses on using deep learning to integrate prior knowledge from a training dataset~\cite{mayer2016large,laga2020survey}. While these approaches need heavy computing power and perform very well on some datasets, they suffer from robustness and limited generalizability~\cite{yamanaka2020adversarial,song2021adastereo}. In contrast, our method is explicit, does not need training data, and uses a quantum computer instead of graphics processing units. 

For traditional approaches, Scharstein \etal ~\cite{middlebury2001} presented a taxonomy  where they introduced the building blocks (1) matching cost computation, (2) cost (support) aggregation, (3) disparity optimization, and (4) disparity refinement. The focus of our work is on (3) as we present a novel way to implement the optimization with a QUBO. For the other building blocks, we follow the baseline~\cite{9653310}. When seeing stereo matching as an optimization problem, it is most common to establish a global energy formulation with a data and smoothness term. In the case of linear regularizers, max-flow-min-cut~\cite{roy1998maximum} methods were shown to be efficient and are also the choice of the previous work for quantum annealers~\cite{9653310}.

However, when considering the  general case and moving to the more robust nonlinear regularizers, the problem becomes {\NPhard} and many heuristics and algorithms to find a good local minimum have been proposed, including continuation~\cite{blake1987visual}, simulated annealing~\cite{article,barnard1989stochastic}, highest confidence first~\cite{chou1990theory} and mean-field annealing~\cite{gennert1988brightness}. The most popular traditional techniques that allow leveraging nonlinear regularizers and stand in contrast to ours are belief propagation~\cite{1206509} and semi-global matching~\cite{hirschmuller2005accurate}. However, these require iterative updates and cannot be formulated as a QUBO. Unlike prior work in this area, we use quantum annealers to tackle an {\NPhard} problem. 

\subsection{Quantum Computer Vision} 

Quantum Computer Vision (QCV) is an emerging field and many methods for different problems were proposed over the last years, such as object tracking \cite{zaech2022adiabatic}, robust fitting \cite{Doan2022} and motion segmentation \cite{Arrigoni2022}. Several techniques tackle correspondence problems across two or more instances, \textit{i.e.,} graph and matching \cite{SeelbachBenkner2020}, mesh alignment \cite{SeelbachBenkner2021}, point set registration \cite{Meli_2022_CVPR, NoormandipourWang2022} and stereo matching \cite{CruzSantos2018AQF, 9653310}. 

The method of Heidari \etal ~\cite{9653310} leverages the graph cut formulation introduced by Cruz-Santos \etal \cite{CruzSantos2018AQF}, and is the closest work to ours, but in contrast, does not allow for nonlinear regularizers and is solvable in linear time. The quantum stereo matching method of Heidari \etal~\cite{9653310} is the closest work to ours. It casts stereo matching as a min-flow max-cut problem converted into a QUBO problem using techniques introduced by Cruz-Santos \etal \cite{CruzSantos2018AQF}, which were improved upon by Krauss \etal \cite{9224181}. Alternatively, stereo matching can be formulated as an MRF MAP (\textit{Markov Random Field Max a Posteriori}) inference, and we follow this approach. To this end, we show how an MRF MAP inference can be transformed into a QUBO. Our QUBO requires the same number of binary variables as Heidari \etal when considering the same region and number of disparities. Although the topological complexity of our QUBO will grow faster than in~\cite{9653310} as the number of possible disparities increases, it will still remain relatively sparse. The use of a coarse-to-fine pyramid allows us to fully embed practical problems onto modern quantum hardware, which was not possible in \cite{9653310}. 

Our method stands in contrast to Presles \etal ~\cite{Presles2023}, which solves an MRF MAP via a graph cut. They introduce an ancillary variable $z_\alpha$ which must connect to all original variables. Since $z_\alpha$ will have so many connections, embedding this QUBO problem onto a quantum annealer will be difficult, and this difficultly will scale poorly as the problem grows. Additionally, to enforce that their connected ancillary variable $z_\alpha$ is always $1$, they must set the corresponding weight to be extremely negative. This can cause the annealer's energy landscape to be extremely jagged and lead to lower accuracy. Another method for formulating MRF MAP inferences as QUBOs \cite{OtgonbaatarDatcu2020} is limited to the binary MRF case, which severely limits the scope of the practical problems which their formulation can solve. In contrast, ours is applicable to any label space size. 
\section{Method}

\subsection{Background}\label{subsec:background}

\newcommand{\vx}[0]{\mathbf{x}}
\newcommand{\vv}[0]{\mathbf{v}}
\newcommand{\vp}[0]{\mathbf{p}}
\newcommand{\vq}[0]{\mathbf{q}}
\newcommand{\lab}[1]{\ell_{#1}}

We aim at a formulation compatible with experimental adiabatic quantum annealers such as D-Wave. Unlike gate-based quantum computers, the adiabatic quantum annealers are already suitable for practically relevant combinatorial optimization problems, and provide a speed-up over traditional machines in solving them \cite{PhysRevX.8.031016, YanSinitsyn2022}. D-Wave exclusively supports \textit{Quadratic Unconstrained Binary Optimization} (QUBO) objectives and, hence, all target tasks, including possible boundary conditions, must be converted to a QUBO before quantum annealing can be attempted. 

Let $\vx \in \{0,1\}^n $ be a binary vector of length $n$. A QUBO problem is defined as finding $\vx^*$ such that: 
\begin{equation}\label{eq:QUBO_def}
    \vx^* = \text{argmin}_{\vx \in \{0,1\}^n} \vx^T Q \vx\,\,, 
\end{equation}
where $Q \in \mathbb{R}^{n \times n}$ is symmetric. $\vx^TQ\vx$ is the quadratic form of $Q$, \textit{i.e.,} it is a polynomial in terms of the entries of $\vx$ of at most degree $2$. 

Quantum annealers solve \eqref{eq:QUBO_def} by mapping the QUBO onto a quantum-mechanical system consisting of qubits. A qubit is an object small enough to have its behavior be governed by quantum mechanics. When we measure a qubit, we will observe it to be either the state $\ket{0}$ or $\ket{1}$. The possible energies of the annealer's system are described by a quantum mechanical operator called the Hamiltonian, $\mathcal{H}_P$. By measuring the state of every qubit in the system, and by referring to $\mathcal{H}_P$, we can determine the total energy of the system. In our mapping, each binary variable in $\vx$ is mapped to a qubit, and $Q$ is mapped to $\mathcal{H}_P$ such that when the qubits are measured, the system's total energy is equal to $\vx^TQ\vx$. We seek to measure the ground (\textit{i.e.,} lowest-energy) state of the system as permitted by $\mathcal{H}_P$, as this is equivalent to finding $\vx^*$. Even if we cannot measure the ground state, any low energy state should be a reasonably close solution to the QUBO.

In these terms, quantum annealing (QA) works as follows: The annealer initializes with possible energies described by a standard initial Hamiltonian $\mathcal{H}_I$, and with qubits in the known ground state. Next, during annealing, the system smoothly transitions from being described by $\mathcal{H}_I$ to $\mathcal{H}_P$. The adiabatic theorem of quantum mechanics \cite{BornFock1928} states that if the interpolation between $\mathcal{H}_I$ and $\mathcal{H}_P$ is slow enough, the system will have a non-zero (and often high) probability to remain in its ground state. After annealing, the state of the system is measured. Ideally, this is the ground state of $\mathcal{H}_P$, however it might only be close to the ground state. Annealing is often run many times, with the measured state with the minimal energy being returned as the proposed QUBO solution.
 
Further details on qubit measurement and the speed ranges of $\mathcal{H}_I$ to $\mathcal{H}_P$ transitions can be found in Nielsen and Chuang~\cite{NielsenChuang2011} and the works by Farhi~et~al.~\cite{Farhi2000QuantumCB, Farhi2001}. However, for the remainder of this paper, it is only important to understand that our problem needs to be formulated as a QUBO \eqref{eq:QUBO_def}.

\subsection{Formulating MRF MAP as a QUBO}\label{subsec:formulating_mrfs_as_qubos}
Following the notation in Drory \etal ~\cite{Drory2014SemiGlobalMA}, a \textit{Markov Random Field} (MRF) can be formulated as an undirected graph 
$\mathcal{G} = (\mathcal{V},\mathcal{E})$, 
where each vertex $\vv \in \mathcal{V}$ has a label $\lab{\vv}$ from a discrete set $\mathcal{L}_\vv$, 
and there are unary costs $\varphi_\vv(\lab{\vv})$ and binary costs $\varphi_{\vp, \vq}(\lab{\vp},\lab{\vq})$
for $(\mathbf{p}, \mathbf{q}) \in \mathcal{E}$. The energy of the MRF is then defined as: 
\begin{equation} \label{eq:MRF_energy}
    E(\lab{(\cdot)}) =  \sum_{\vv \in \mathcal{V}} \varphi_\vv(\lab{\vv})
    + \sum_{(\vp,\vq) \in \mathcal{E}} 
    \varphi_{\vp,\vq}(\lab{\vp},\lab{\vq})\,\,. 
\end{equation}
Finding the labelling $\lab{(\cdot)}$ such that $E(\lab{(\cdot)})$ is minimal 
is the MRF \textit{maximum a posteriori} (\textit{MAP}) inference problem, which is {\NPhard} in general~\cite[pg.551]{koller2009probabilistic}. This motivates us to use adiabatic quantum computing and in this subsection, we will show how such a general MRF can be mapped to a QUBO. Subsequently, we will show how stereo matching can be formulated as an instance of an MRF and solved in a quantum-hybrid manner. Notably, the QUBO formulation of an MRF is general and could be applied to a wide range of other problems in the future. 

A \emph{binary} encoding scheme for our Markov variable labels (to be encoded into a QUBO) is conceivable (see \cref{sec:binary_encoding_of_mrf_map}); this approach would avoid introducing rectifiers necessary with our other scheme and, therefore, may increase the annealing stability. Unfortunately, the number of QUBO binary variables required to represent the MRF in this case grows exponentially with the number of labels, which  quickly becomes infeasible with current and likely near-term future hardware. For a full analysis on embedding complexities of all schemes, see \cref{sec:embedding_problem_graphs}. 

Therefore, we proceed with a \emph{one-hot} encoding scheme of the Markov variable labels together with a novel local rectifier that minimizes the disturbances during annealing. For a given vertex $\vv$, the set of labels is denoted by $\mathcal{L}_\vv$ and is enumerated by the index $l$. For each possible label value $\lab{\vv}^l$, we create a corresponding binary variable $\vx_{\lab{_\textbf{v}}^l}$:
\begin{equation} \label{eq:one_hot_binary_variable_meaning}
    \vx_{\lab{\vv}^l}=
    \begin{cases}
        1 & \text{if } \lab{\vv} = \lab{\vv}^l\,\,,\\
        0 & \text{else\,\,.} 
    \end{cases}
\end{equation}
This allows us to rewrite~\cref{eq:MRF_energy} as:
\begin{equation} \label{eq:markov_cost_quadratic_form}
    \begin{split}
    E(\lab{(\cdot)}) &= \sum_{\vv \in \mathcal{V}} \, \sum_{\lab{\vv}^l \in \mathcal{L}_\vv} \varphi_\vv (\lab{\vv}^l)\vx_{\lab{\vv}^l}^2 + \\
    &+ \sum_{(\vp,\vq) \in \mathcal{E}} \, \sum_{\ell_\vp^r \in \mathcal{L}_\vp} \, \sum_{\ell_\vq^s \in \mathcal{L}_\vq} \varphi_{\vp,\vq}(\ell_\vp^r, \ell_\vq^s)\vx_{\ell_\vp^r}\vx_{\ell_\vq^s} \,\,.
    \end{split} 
\end{equation}
Next, we can write our Markov random field cost as a quadratic polynomial in terms of a collection of binary variables. To bring the cost from~\cref{eq:markov_cost_quadratic_form} into the  quadratic form~\eqref{eq:QUBO_def}, we define the binary vector $\vx$ as 
\begin{equation} \label{eq:x_full_vector_definition}
    \vx = \{\vx_{\lab{\vv}^l}\}_{\forall \vv \in \mathcal{V},\, \forall \lab{\vv}^l \in \mathcal{L}_\vv } \,\,. 
\end{equation}
We index entries of $Q$ by $\lab{\vp}^r, \lab{\vq}^s$ and define them as follows: 
\begin{equation}\label{eq:markov_qubo_matrix_no_constraints}
\small
\begin{split}
    \forall \vv \in \mathcal{V}, \, \forall \lab{\vv}^l \in \mathcal{L}_\vv: \,\, Q_{\lab{\vv}^l,\lab{\vv}^l} & = \varphi_\vv(\lab{\vv}^l) \\
    \forall (\vp,\vq) \in \mathcal{E},\, \forall \lab{\vp}^r \in \mathcal{L}_\vp,\, \forall \lab{\vq}^s \in \mathcal{L}_\vq: \,\, 
    Q_{\lab{\vp}^r, \lab{\vq}^s} & = \frac{1}{2}\varphi_{\vp,\vq}(\lab{\vp}^r,\lab{\vq}^s) \\ 
       \,\, Q_{\lab{\vq}^s, \lab{\vp}^r} & = \frac{1}{2}\varphi_{\vq,\vp}(\lab{\vq}^s,\lab{\vp}^r) \,\,. 
\end{split}
\end{equation}
All other entries of $Q$ are $0$. We have now translated our label from~\cref{eq:markov_cost_quadratic_form} into the matrix quadratic form:  
\begin{equation} \label{eq:quadratic_form_equals_markov_energy}
    \vx^TQ\vx = E(\lab{(\cdot)})\,\,. 
\end{equation}
Note, however, that we cannot submit $Q$ to a quantum annealer in its current form as we must enforce that for every vertex we have only one label assigned, \textit{i.e.,} there must be exactly one binary variable $x_{\lab{\vv}^l}$ that is equal to $1$. We can express this constraint as:
\begin{equation}\label{eq:onehot_qubo_constraints}
    \forall \vv \in \mathcal{V}: \sum_{\forall \lab{\vv}^l \in \mathcal{L}_\vv} \vx_{\lab{\vp}^l} = 1\,\,. 
\end{equation}
Following the techniques proposed in QSync~\cite{birdal2021quantum}, we can incorporate this constraint by tweaking our definition of $Q$ from~\cref{eq:markov_qubo_matrix_no_constraints} into the following form: 
\begin{equation}
\small
\begin{split}
    \forall \vv \in \mathcal{V},\, \forall \lab{\vv}^l \in \mathcal{L}_\vv: & \,\, Q_{\lab{\vv}^l,\lab{\vv}^l} = \varphi_\vv(\lab{\vv}^l)  - \Lambda(\lab{\vv}^l,\lab{\vv}^l) \\
    \forall \vv \in \mathcal{V},\, \forall \lab{\vv}^r,\lab{\vv}^s \in \mathcal{L}_\vv,\, r \neq s: & \,\, Q_{\lab{\vv}^r,\lab{\vv}^s} = \Lambda(\lab{\vv}^r,\lab{\vv}^s) \\
    & \,\, Q_{\ell_\vv^s,\ell_\vv^r} = \Lambda(\lab{\vv}^r,\lab{\vv}^s) \\
    \forall (\vp,\vq) \in E, \forall \ell_\vp^r \in \mathcal{L}_\vp \forall \ell_\vq^s \in \mathcal{L}_\vq: & \,\, Q_{\ell_\vp^r, \ell_\vq^s} = \frac{1}{2}\varphi_{\vp,\vq}(\ell_\vp^r,\ell_\vq^s) \\ 
      & \,\, Q_{ \ell_\vq^s, \ell_\vp^r} = \frac{1}{2}\varphi_{\vp,\vq}(\ell_\vp^r,\ell_\vq^s), \\
\end{split}
\label{eq:quadratic_form_equals_markov_energy_with_granular_constraints}
\end{equation}
with the unmentioned entries of $Q$ being $0$. The common rectification scheme is to set
$\Lambda(\lab{\vv}^r,\lab{\vv}^s)=\lambda_\vv$, with $\lambda_\vv$ being a vertex-specific constant which enforces that only one label per vertex should have a coefficient of $1$. However, a higher $\Lambda$ leads to a more jagged energy landscape on the quantum annealer and negatively affects its results. Thus, one would like to set the $\Lambda$ as small as possible (see~\cref{sec:rectifier_strength} for details). To this end, we derive a function $\Lambda(\cdot, \cdot)$ that yields a sufficiently high upper bound while obeying our constraints. The full derivation of $\Lambda(\cdot, \cdot)$ is presented in~\cref{sec:deriving_rectifiers}. 

\subsection{Stereo Matching as an MRF MAP}\label{subsec:stereo_matching_as_an_mrf}
Let $\{I^L, I^R\}$ be a set of grayscale stereo images defined over the pixel grid domain $\Omega \subset \mathbb{N}^2$. We assume that our images are rectified, meaning that the per-point displacements in the image plane lie on horizontal epipolar lines and are positive. We denote the disparity at image coordinates $(i,j)$ with $d_{i,j}$ and the matrix containing all disparities as $D$. There are many works that treat stereo matching as an energy minimization problem~\cite{9653310,xue2016stereo,7018068}, where the energy is the sum of a data term $E_d(d_{i,j})$ across $D$ and a smoothness term $E_s(d_{i,j}, d_{i',j'})$  with $(i,j)$ and $(i',j')$ being from the set $\mathcal{N}$ of neighboring pixels. The data term should be lower when the estimated disparities map to similar regions in the second image. In our case, it relies on the brightness constancy assumption  \cite{Gennert1988BrightnessbasedSM, 9653310}: 
\begin{equation}
    E_d(d_{i,j}) = (I^L(i,j) - I^R(i - d_{i,j}, j))^2\,.
    \label{eq:brightness_constancy_assumption}
\end{equation}
The smoothness term should be lower when the disparities maintain local structural coherence. We use a truncated (nonlinear) regularizer with edge-awareness: 
\begin{equation}
    E_s(d_{i,j}, d_{i',j'}) = \begin{cases}
 R \text{    if } |I^L(i,j) - I^L(i',j') | \leq \tau \\
 R / q \text{ else } 
\end{cases},
    \label{eq:edge_aware_potts_model}
\end{equation}
with 
\begin{equation}
    R = \operatorname{min}(m, s\,|d_{i,j} - d_{i',j'}|), 
    \label{eq:truncated_linear_regularizer}
\end{equation}
where $\tau$, $q$, $m$ and $s$ are tunable hyperparameters. Similar ideas for nonlinear, edge aware regularizers were discussed in the literature before \cite{Scharstein2001}. This nonlinear regularizer makes the resulting total energy function (See \cref{eq:stereo_matching_energy_function} \NPhard{} to minimize. At the same time, this regularizer has only a few hyperparameters, making it relatively simple to tune and optimize. Throughout \cref{sec:results}, we fix our regularization parameters across all experiments. The values of all the parameters at each iteration of our coarse-to-fine algorithm are given in \cref{sec:regularization_hyper_parameters}. 

We can now define our total energy functional $E(D)$ as
\begin{equation}
  E(D) = \sum_{(i,j) \in \Omega} E_d(d_{i,j}) + \sum_{((i,j),(i'j')) \in \mathcal{N}} E_s(d_{i,j}, d_{i',j'})\,. 
  \label{eq:stereo_matching_energy_function}
\end{equation}
We seek the disparity matrix $D^*$ that minimizes $E(D)$. Notably, the transformation into an MRF is straight forward with 
\begin{equation}
\small
\begin{split}
    &\mathcal{V} =  \Omega\,,\, 
    \mathcal{E} =  \mathcal{N}\,,\, 
    \lab{i,j} =  d_{i,j} \,,
    \varphi_{(i,j)}(\lab{i,j}) = E_d(d_{i,j}) \,\, \text{and}\,\\
    &\varphi_{(i,j),(i',j')}(\lab{i,j}, \lab{i',j'}) = E_s(d_{i,j}, d_{i',j'}) \,.\\
\end{split}
\end{equation}
\subsection{Our Stereo Matching Algorithm} \label{subsec:our_algorithm}
We now describe our full method for stereo matching that leverages adiabatic quantum computing. When given an image pair, we first precalculate $E_d(d_{i,j})$ and $E_s(d_{i,j}, d_{i',j'})$ for all considered disparity sizes on a traditional machine. Following the derivation in the previous sections, we then produce the matrix $Q$, which we can submit to either a traditional QUBO solver or a quantum annealer. In either case, we receive the binary vector $\vx^*$ as a response that we  decode into the disparity matrix $D^*$. 

As the QUBO formulations of stereo estimation problems can become too large to be mapped to quantum hardware or even solved directly with a traditional optimizer, we employ a coarse-to-fine strategy. First, we downsample the images by a factor of $4$ in each dimension, and solve the QUBO for $6$ possible disparities per pixel, corresponding to disparities of $0$, $4$, $8$, $12$, $16$, and $20$ on the full resolution. We then upsample by $2$ to proceed to the next higher resolution, this time considering $4$ possible labels. Finally, we upsample to full resolution and consider $4$ possible labels. This method allows us to consider a sufficient number of disparity levels arising in stereo matching problems\footnote{\textit{e.g.,} all disparity levels present in the Middlebury dataset \cite{middlebury2001}}. The method can, in theory, be just as accurate with only two coarse-to-fine iterations, one at the downsample factor of $4$, and one at full resolution. However, we found that including intermediate coarseness layers allows for some ability to correct errors and yields the best results (see ~\cref{sec:coarse_to_fine_ablation}).

To reduce the QUBO size further, we split the problem into \textit{bundles} of epipolar lines and solve for each bundle individually. Similar to Heidari \etal ~\cite{9653310}, we observe that the solutions obtained by our method are still noisy. Therefore, we follow their denoising approach and apply median filtering to our disparity estimate at each resolution before upsampling, and bilateral filtering \cite{710815} as a final post-processing step. We have now obtained a quantum-hybrid method for stereo matching that can leverage modern quantum hardware for combinatorial optimization objectives. A full overview of the method is given in Algorithms \ref{alg:full_quantum_algorithm} and \ref{alg:solve_mrf_map}. 

 \begin{algorithm}
 \small
  \caption{Coarse-to-Fine Stereo Matching}\label{alg:full_quantum_algorithm}
  \begin{algorithmic}[1]
    \Procedure{Stereo Match}{$I^L,I^R$}
      \State $\textit{resolutions} \gets [\frac{1}{4},\frac{1}{2},1]$
      \State $\textit{disparity ranges} \gets [6,4,4]$
      \State $\mathcal{D}^* \gets 0$
      \For{\texttt{step} in $[1,2,3]$}
        \State $\textit{r} \gets \textit{resolutions}[\texttt{step}]$
        \State $\textit{dr} \gets \textit{disparity ranges}[\texttt{step}]$
        \State $\textit{pd} \gets \Call{Disparity Ranges At Res}{r,dr,\mathcal{D}^*}$ 
        \State $I^L_\texttt{step} \gets \Call{Resize}{I^L,r}$
        \State $I^R_\texttt{step} \gets \Call{Resize}{I^R,r}$
        \State $E_d \gets \Call{Build Data Terms}{pd, I^L_\texttt{step},I^R_\texttt{step}}$
        \State $E_s \gets \Call{Build Smoothness Terms}{pd, I^L_\texttt{step},I^R_\texttt{step}}$
        \State $\varphi_\vp \gets E_d$ \Comment{Convert to MRF notation}
        \State $\varphi_{\vp,\vq} \gets E_s$ \Comment{Convert to MRF notation}
        \State $\ell^* \gets \Call{Solve MRF MAP}{\varphi_\vp,\varphi_{\vp,\vq} }$ (\text{\cref{alg:solve_mrf_map}}) 
        \State $\mathcal{D}^* \gets \ell^*$ \Comment{Convert from MRF notation}
        \State $\mathcal{D}^* \gets \Call{Full Size From Resolution}{\mathcal{D}^*, r}$
        \State $\mathcal{D}^* \gets \Call{Median Filter}{\mathcal{D}^*}$
      \EndFor
      \State $\mathcal{D}^* \gets \Call{Bilateral Filter}{\mathcal{D}^*}$
      \State \textbf{return} $\mathcal{D}^*$
    \EndProcedure
  \end{algorithmic}
\end{algorithm}
\begin{algorithm}
  \caption{MRF MAP Solver via QUBO}\label{alg:solve_mrf_map}
  \begin{algorithmic}[1]
    \Procedure{Solve MRF MAP}{$\varphi_\vp,\varphi_{\vp,\vq} $}
      \State $Q \gets \Call{Encode QUBO}{\varphi_\vp,\varphi_{\vp,\vq}}$ (\cref{eq:quadratic_form_equals_markov_energy_with_granular_constraints}) 
      \State $x^* \gets \Call{Anneal}{Q}$ \Comment{D-Wave or trad. solver}
      \State $\ell^* \gets \Call{Decode Annealer Response}{x^*}$ 
      \State \textbf{return} $\ell^*$
    \EndProcedure
  \end{algorithmic}
\end{algorithm}

\section{Experimental Results}\label{sec:results} 

For the following sections, we follow Heidari \textit{et al.}~\cite{9653310} and test on four stereo matching pairs from the Middlebury 2001 dataset~\cite{middlebury2001}: Tsukuba, Bull, Venus, and Sawtooth. We use the root mean squared error (RMSE) and bad pixel percentage (BPP)~\cite{middlebury2001} with $\delta_d := 1$ for numerical evaluations. 

\subsection{Our Experimental Setting}\label{subsec:our_experimental_setting} 
There are many traditional methods to solve QUBO problems. For non-quantum methods, we selected Gurobi~\cite{Gurobi2023} and simulated annealing~\cite{Kirkpatrick1983}. Gurobi uses branch and bound optimization to find a solution near the global optimum within a tight margin. Its output can be viewed as what an ideal quantum annealer would produce. Simulated annealing is a traditional global optimization technique to solve non-convex problems and can be seen as a simulation of a thermal annealing process. We use D-Wave's simulated annealer with default parameters \cite{DWave_Neal}. 

To test with quantum annealing, we use D-Wave's Pegasus QPU with ${\sim}5.4 \cdot 10^4$ qubits \cite{2020arXiv200300133B}. To compute the minor embeddings onto Pegasus, we use D-Wave's minorminer, which by default relies on Cai \textit{et al.}'s algorithm \cite{cai2014practical}. Minor embedding is required as physical qubits have limited connectivity, and the logical (or analytically derived qubits) often need to be mapped to chains of physical qubits to enable sampling of a given problem. A minor embedding is a mapping of QUBO binary variables to specific hardware qubits on the QPU. Since physical qubits are connected in a pre-defined pattern (with limited connectivity), a single binary variable must be often mapped to chains of qubits to represent an arbitrary QUBO correctly. Lastly, we tested our approach on closed-source D-Wave's hybrid solver, which uses traditional optimization in conjunction with quantum annealing. For all techniques involving quantum annealers or simulated annealing, we allow $500$ annealing runs and take the lowest energy solution. All other settings involving D-Wave's API are set to the default (\textit{i.e.,} annealing time of $20\mu$sec, a majority voting policy for resolving broken chains of physical qubits, and a chain strength calculated 
as $C = 1.414 R \sqrt{D}$, with $R$ being the standard deviation of quadratic QUBO coefficients and $D$ being the average degree of QUBO problem nodes). 

\begin{figure}
    \centering
    \setlength\tabcolsep{1.5pt}
    \begin{tabular}{c c c c c}
                   & Tsukuba & Bull & Sawtooth & Venus \\
        \rotatebox{90}{\textcolor{white}{------}\textbf{LI}} &\includegraphics[width=\tableimagewidth, height=\tableimageheight]{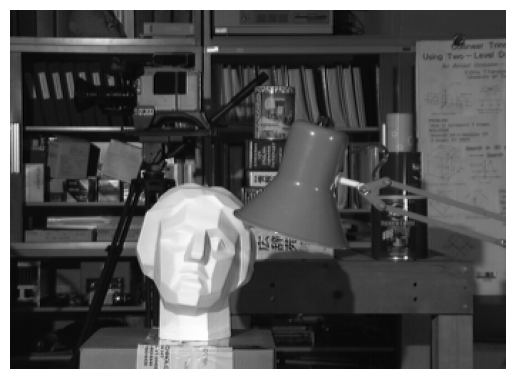} & \includegraphics[width=\tableimagewidth, height=\tableimageheight]{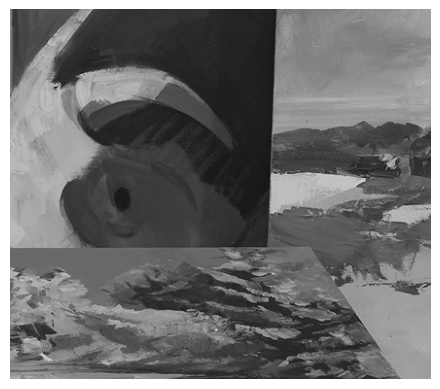}  & \includegraphics[width=\tableimagewidth, height=\tableimageheight]{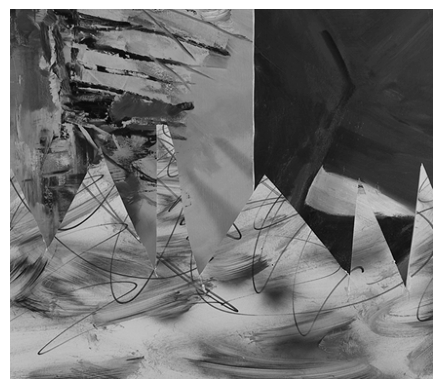} & \includegraphics[width=\tableimagewidth, height=\tableimageheight]{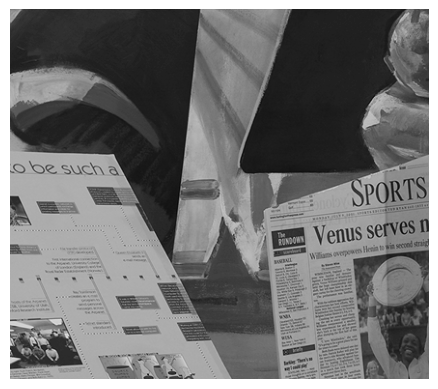} \\
       \rotatebox{90}{\textcolor{white}{-----}\textbf{GT}} & \includegraphics[width=\tableimagewidth, height=\tableimageheight]{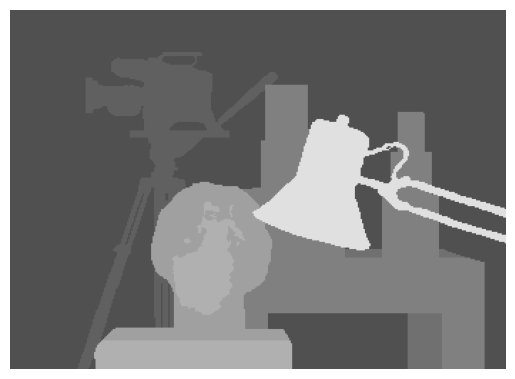} & \includegraphics[width=\tableimagewidth, height=\tableimageheight]{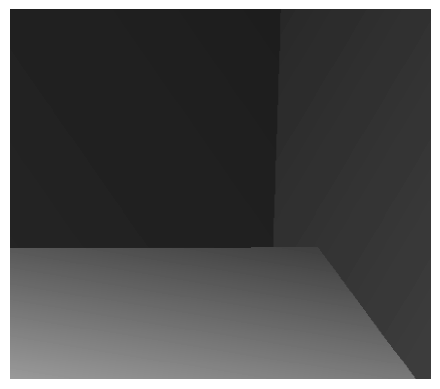}  & \includegraphics[width=\tableimagewidth, height=\tableimageheight]{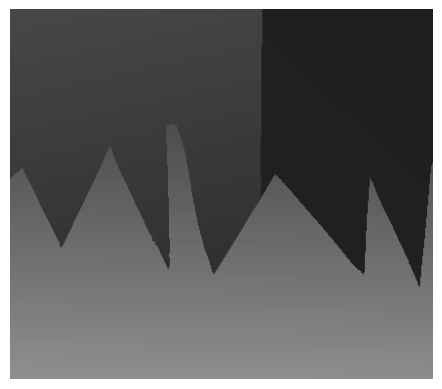} & \includegraphics[width=\tableimagewidth, height=\tableimageheight]{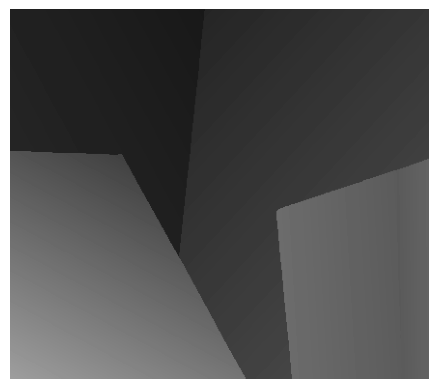}\\
       \rotatebox{90}{\textcolor{white}{--}Ours (\textbf{G})} & \includegraphics[width=\tableimagewidth, height=\tableimageheight]{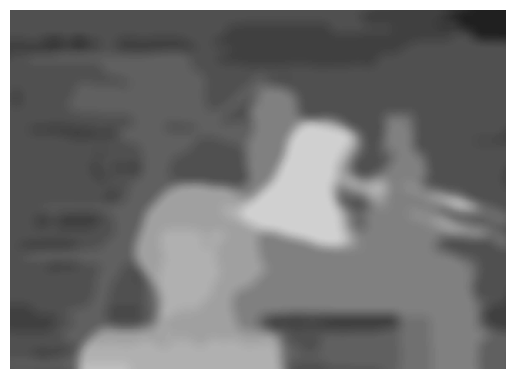} & \includegraphics[width=\tableimagewidth, height=\tableimageheight]{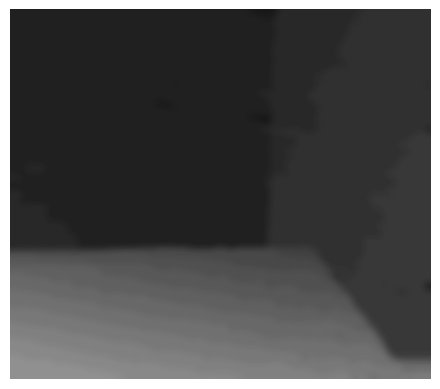}  & \includegraphics[width=\tableimagewidth, height=\tableimageheight]{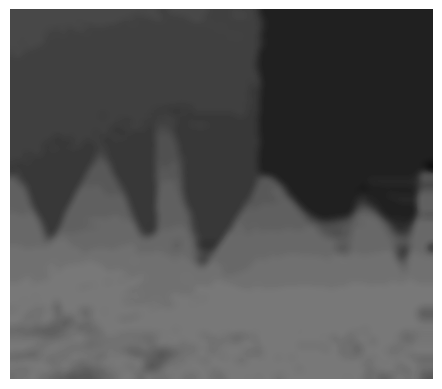} & \includegraphics[width=\tableimagewidth, height=\tableimageheight]{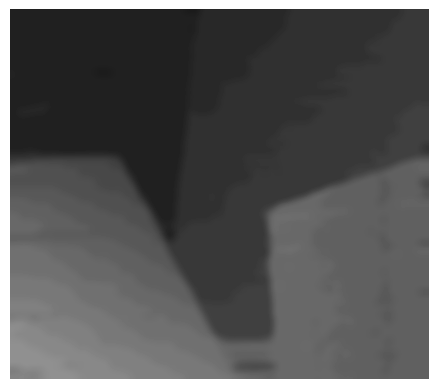}\\
       \rotatebox{90}{\textcolor{white}{-}Ours (\textbf{SA})} & \includegraphics[width=\tableimagewidth, height=\tableimageheight]{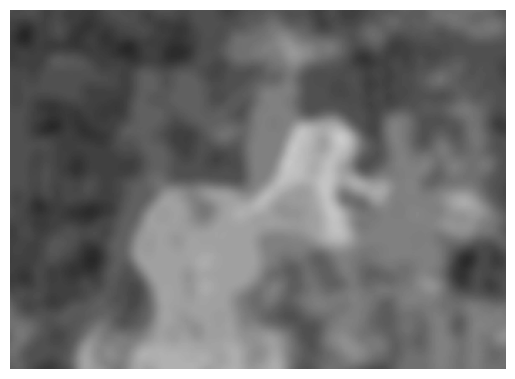} & \includegraphics[width=\tableimagewidth, height=\tableimageheight]{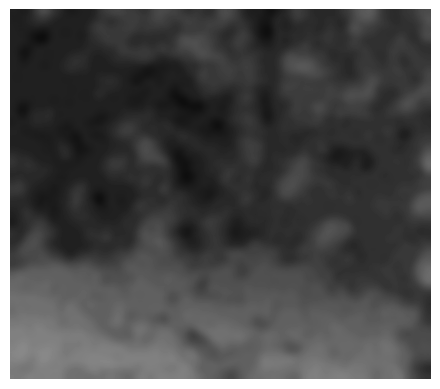}  & \includegraphics[width=\tableimagewidth, height=\tableimageheight]{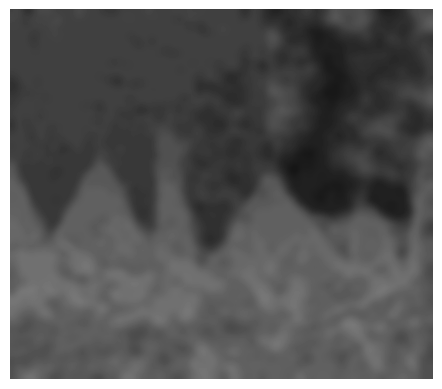} & \includegraphics[width=\tableimagewidth, height=\tableimageheight]{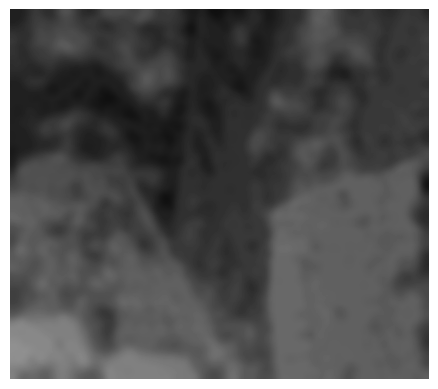}\\
        \rotatebox{90}{\textcolor{white}{--}Ours (\textbf{H})} & \includegraphics[width=\tableimagewidth, height=\tableimageheight]{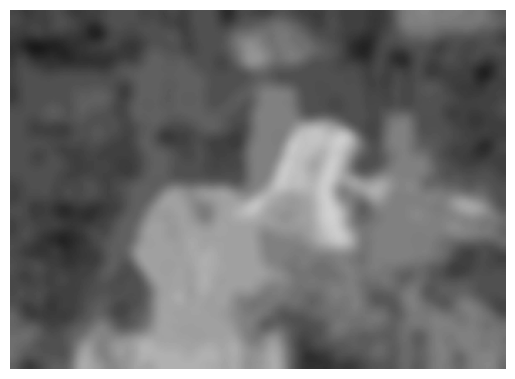} & \includegraphics[width=\tableimagewidth, height=\tableimageheight]{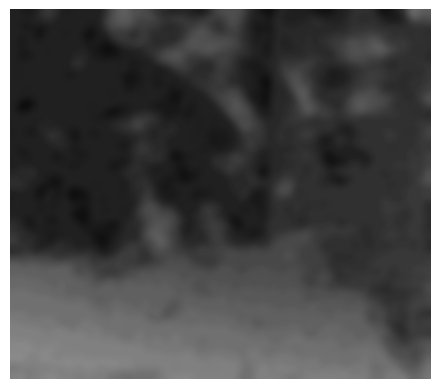}  & \includegraphics[width=\tableimagewidth, height=\tableimageheight]{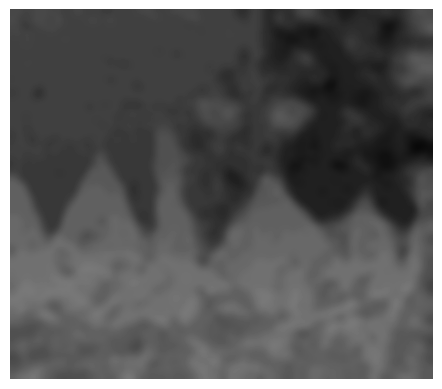} & \includegraphics[width=\tableimagewidth, height=\tableimageheight]{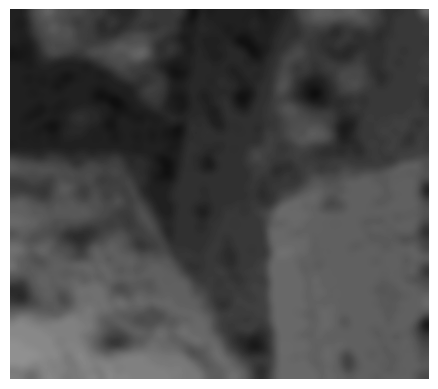}\\
       \rotatebox{90}{\textcolor{white}{--}Ours (\textbf{Q})} & \includegraphics[width=\tableimagewidth, height=\tableimageheight]{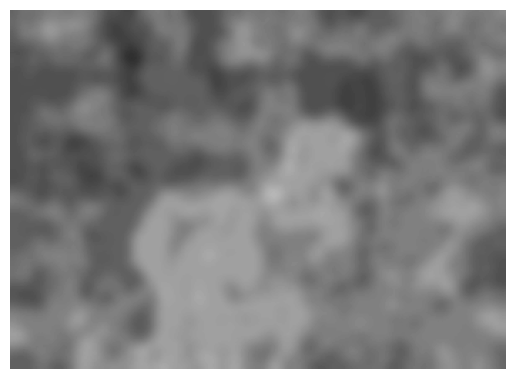} & \includegraphics[width=\tableimagewidth, height=\tableimageheight]{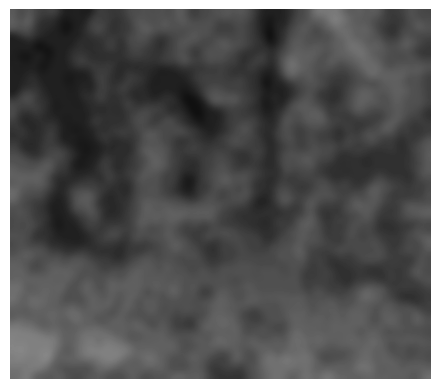}  & \includegraphics[width=\tableimagewidth, height=\tableimageheight]{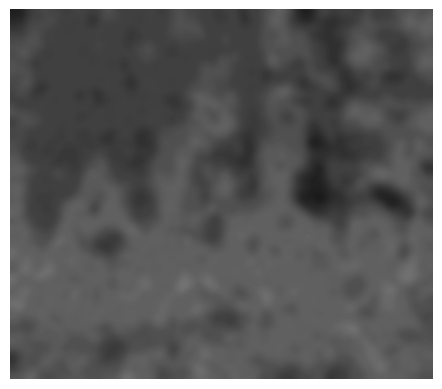} & \includegraphics[width=\tableimagewidth, height=\tableimageheight]{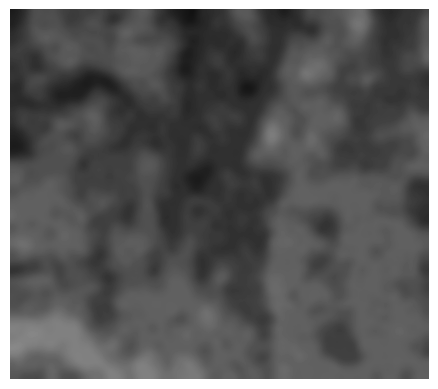}\\
    \end{tabular}
    \caption{The first row shows the \textbf{L}eft \textbf{I}mage from each of the four Middlebury stereo pairs. The second row shows the \textbf{G}round-\textbf{T}ruth displacements for each pair. The remaining rows show results by our method using different optimizers: \textbf{G}urobi, \textbf{S}imulated \textbf{A}nnealing, D-Wave's \textbf{H}ybrid Quantum-Classical Solver, and the D-Wave's Pegasus \textbf{Q}PU. The choice of optimizer has a strong influence on the result quality, and we observe that the traditional optimizer Gurobi outperforms all other methods. We hypothesize that this is because of the jagged and challenging energy landscape for the tested quantum annealer caused by our rectifiers and the current state of quantum hardware.
    } 
    \label{fig:optimizer_comparison}
\end{figure}
For the experiments, we restrict our bundles to be a single epipolar line. First, this makes our results more comparable to the previous quantum-admissible stereo matching approach \cite{9653310} that also operates solely on single epipolar lines. Second, a single epipolar line problem from the Middlebury dataset is sufficiently small to embed onto current D-Wave hardware at all three coarseness levels. We provide a visualization of the different results in \cref{fig:optimizer_comparison} and numerical results in \cref{tab:optimizer_comparison}. 

Our method performs best when using Gurobi. We conjecture that this is because introducing constraints into a QUBO (\textit{i.e.,} rectification of the QUBO data term) can disrupt the energy landscape and impede the annealing process. For a deeper discussion of this phenomenon, we refer to Birdal and Golyanik \textit{et al.}~\cite{birdal2021quantum}. Note that future annealers are expected to improve their properties in sampling of more and more challenging energy landscapes, and achieve performance similar to and even better than Gurobi.  

\begin{figure}
    \centering
     \setlength\tabcolsep{1.5pt}
    \begin{tabular}{c c c}
                   & Pre-Median Filtering & Post-Median Filtering \\
        \rotatebox{90}{\textcolor{white}{-----}$\frac{1}{4}$\, Resolution} &\includegraphics[width=\ctoftableimagewidth, height=\ctoftableimageheight]{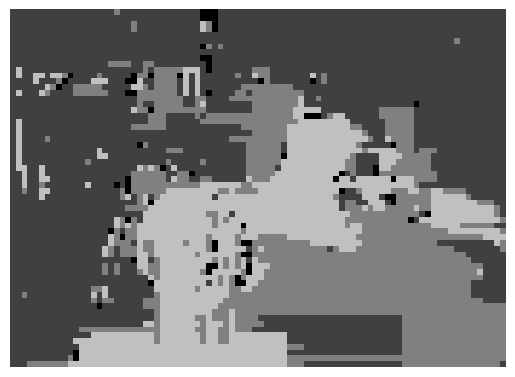} & \includegraphics[width=\ctoftableimagewidth, height=\ctoftableimageheight]{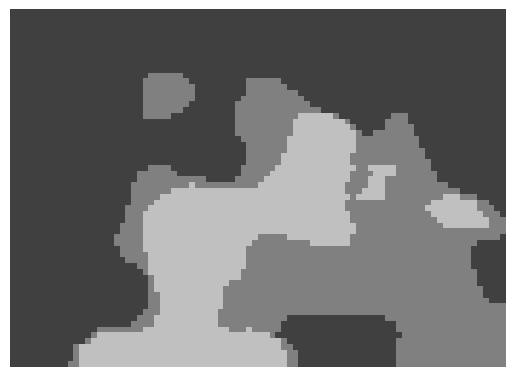}  \\
       \rotatebox{90}{\textcolor{white}{-----}$\frac{1}{2}$\, Resolution} &\includegraphics[width=\ctoftableimagewidth, height=\ctoftableimageheight]{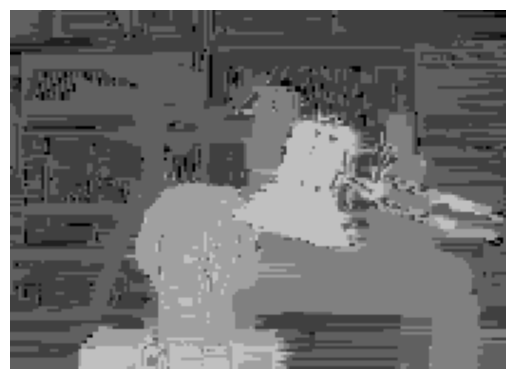} & \includegraphics[width=\ctoftableimagewidth, height=\ctoftableimageheight]{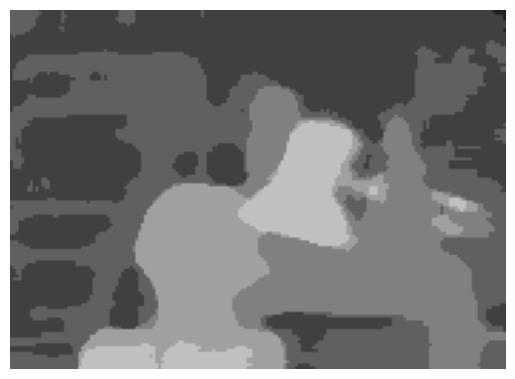}  \\
       \rotatebox{90}{\textcolor{white}{-----}Full Resolution} &\includegraphics[width=\ctoftableimagewidth, height=\ctoftableimageheight]{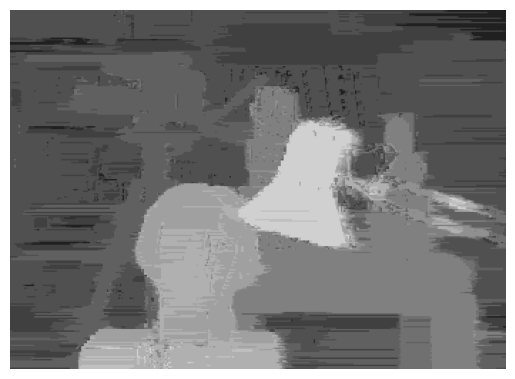} & \includegraphics[width=\ctoftableimagewidth, height=\ctoftableimageheight]{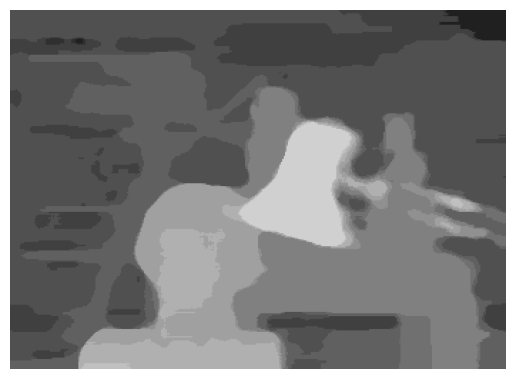}  \\
    \end{tabular}
    \caption{Stereo estimation on the Tsukuba image pair at the three resolution levels, before and after median filtering using the Gurobi solver. Median filtering helps to prevent cascading errors from lower resolution estimates. The final result after bilateral filtering is shown in \cref{fig:optimizer_comparison}.}
    \label{fig:coarse_to_fine_vis}
\end{figure}

Due to the success of the Gurobi optimizations, for subsequent results, we use it unless stated otherwise. In \cref{fig:coarse_to_fine_vis}, we visualize the coarse-to-fine steps of our method. Similar to the previous quantum stereo matching approach \cite{9653310}, the initial results are noisy, while applying median filtering is able to remove the noise effectively. In all the annealing-based methods, we observe splotchy artifacts, which come from coarser resolutions of the pyramid and represent a drawback of coarse-to-fine methods: even with intermediate filtering for corrections, small errors can compound.

In the Gurobi case, these splotchy artifacts are negligible. However, some minor artifacts are still apparent. The streaking patterns (particularly clear on Tsukuba) are likely to occur when one optimizes the epipolar lines independently. We can also see such artifacts in Heidari \etal \cite{9653310}. The Sawtooth example has two additional unwanted artifacts: over-regularization along the jagged edges, and some general inaccuracies in the lower left region. The over-regularization can be attributed to an incompatible parameterization of our regularizer. We have to select hyperparameters which work well in general, and these jagged edges are a corner case. We conjecture that the poor estimation in the lower left region is also due to our energy model parameterization. The images have a lot of texture in that region of the frame which is misinterpreted as edges of objects, causing under regularization. Additionally, the texture contains repeating brightnesses across the epipolar line, weakening our brightness constancy assumption. 

\begin{table}[]
    \centering
    \begin{adjustbox}{width=\columnwidth,center}
    \begin{tabular}{|c| c | c || c | c || c | c || c | c |}
        \hline
        \multirow{2}{*}{\textbf{Image Pair}} &  \multicolumn{2}{ c ||}{\textbf{G}urobi} & \multicolumn{2}{ c ||}{\textbf{S}imulated \textbf{A}nnealing} & \multicolumn{2}{ c ||}{\textbf{H}ybrid} &  \multicolumn{2}{ c |}{\textbf{Q}PU} \\ \cline{2-9}
            & RMSE & BPP & RMSE & BPP & RMSE & BPP & RMSE & BPP \\
            \hline
        \textbf{Tsukuba} & 1.53 & 12.93 & 1.87 & 30.64 & 1.79 & 26.82 & 2.24 & 45.62  \\ \hline
        \textbf{Bull}    & 0.58 & 3.46 & 1.86 & 45.29 & 1.66 & 31.57 & 3.51 & 76.98   \\ \hline
        \textbf{Sawtooth} & 1.89 & 24.51 & 2.27 & 47.73 & 2.71 & 47.49 & 3.99 & 74.24  \\ \hline
        \textbf{Venus}   & 0.96 & 8.16 & 3.04 & 56.75 & 2.17 & 42.43 & 3.24 & 67.70 \\ \hline
        \textbf{Average} & \textbf{1.24} & \textbf{12.27} & 2.26 & 45.10 & 2.08 & 37.08 & 3.25 & 66.14 \\ \hline
        \end{tabular}
    \end{adjustbox}
    \caption{The Root Mean Squared Error (RMSE) and Bad Pixel Percentage (BPP) of our method using all the optimizers used in \cref{fig:optimizer_comparison}. \textbf{G}urobi outperforms all other optimizers on both metrics}
    \label{tab:optimizer_comparison}
\end{table}

\subsection{Comparison to Heidari {\large \textbf{\textit{et al.}}}~\cite{9653310}} 

We compare our results to those of Heidari \etal's \cite{9653310}. However, when making such a comparison, we note that D-Wave's hybrid optimizer is proprietary. Therefore, it is impossible to diagnose to what degree Heidari \etal's QUBO problems were optimized with traditional or quantum hardware. To this end, we compare our method using Gurobi against Heidari \etal's and find that according to~\cref{tab:hybrid_annealing_comp} ours is $22.5\%$ better in RMSE on average, although we do have a higher BPP. This can be considered an upper bound of the performance improvement we can achieve. We also considered how Heidari \etal's results would look if they used classical optimization. For the classical optimization of their formulation, we solved the max-flow min-cut problem using the Ford Fulkerson algorithm \cite{10.5555/1942094}. We do not have access to their regularization weight, and determined a value empirically that we found to yield strong performance. Even in this case, our method has a $2\%$ improvement in RMSE over Heidari \etal. We conclude that the improvement due to our method lies between $2\%$ and  $22.5\%$. 

\begin{figure}
    \centering
     \setlength\tabcolsep{1.5pt}
    \begin{tabular}{c c c c c}
                   & Tsukuba & Bull & Sawtooth & Venus \\
       \rotatebox{90}{ \phantom{ } Ours (\textbf{H})} & \includegraphics[width=\tableimagewidth, height=\tableimageheight]{sec/images/Best_In_Show_Hybrid/tsukuba.png} & \includegraphics[width=\tableimagewidth, height=\tableimageheight]{sec/images/Best_In_Show_Hybrid/bull.png}  & \includegraphics[width=\tableimagewidth, height=\tableimageheight]{sec/images/Best_In_Show_Hybrid/sawtooth.png} & \includegraphics[width=\tableimagewidth, height=\tableimageheight]{sec/images/Best_In_Show_Hybrid/venus.png}\\
       \rotatebox{90}{ Heidari(\textbf{H})} & \includegraphics[width=\heidarihybridtableimagewidth, height=\heidarihybridtableimageheight]{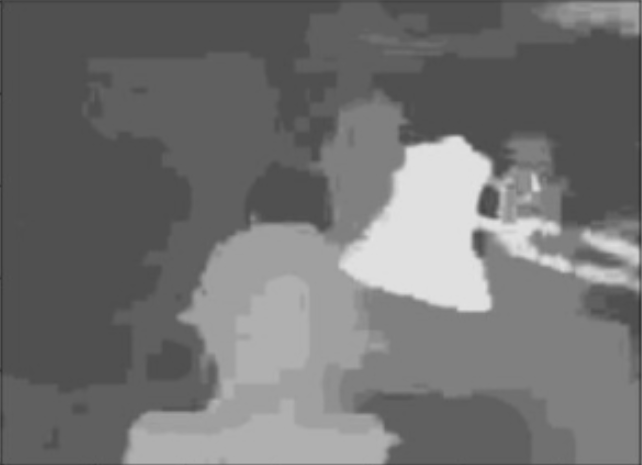} & \includegraphics[width=\heidarihybridtableimagewidth, height=\heidarihybridtableimageheight]{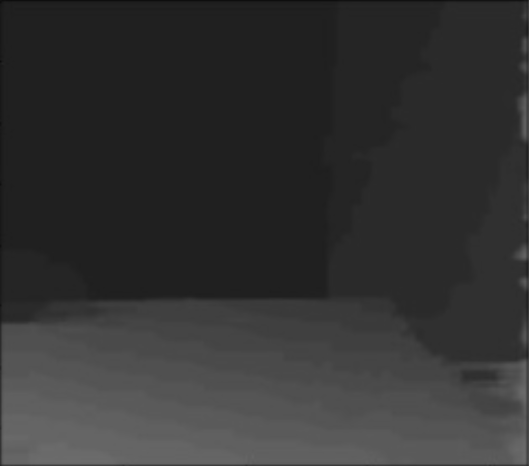}  & \includegraphics[width=\heidarihybridtableimagewidth, height=\heidarihybridtableimageheight]{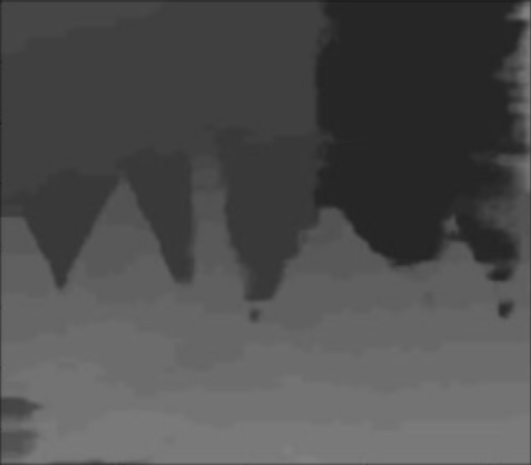} & \includegraphics[width=\heidarihybridtableimagewidth, height=\heidarihybridtableimageheight]{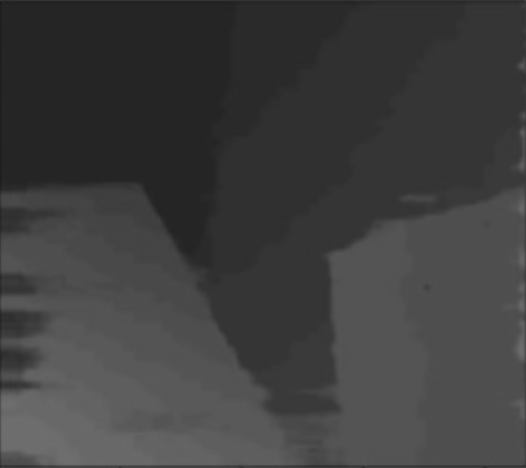}\\
        \rotatebox{90}{\phantom{ } Ours (\textbf{G})} & \includegraphics[width=\tableimagewidth, height=\tableimageheight]{sec/images/Best_In_Show_Gurobi/tsukuba.png} & \includegraphics[width=\tableimagewidth, height=\tableimageheight]{sec/images/Best_In_Show_Gurobi/bull.png}  & \includegraphics[width=\tableimagewidth, height=\tableimageheight]{sec/images/Best_In_Show_Gurobi/sawtooth.png} & \includegraphics[width=\tableimagewidth, height=\tableimageheight]{sec/images/Best_In_Show_Gurobi/venus.png}\\
        \rotatebox{90}{ Heidari (\textbf{F})} & \includegraphics[width=\tableimagewidth, height=\tableimageheight]{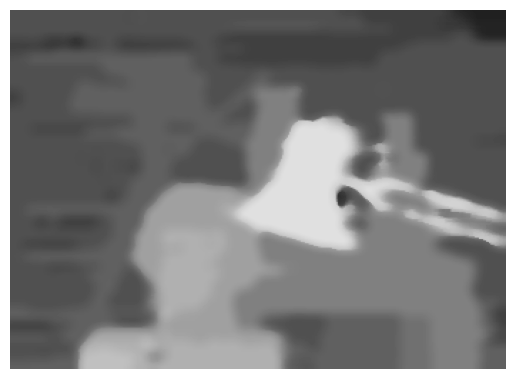} & \includegraphics[width=\tableimagewidth, height=\tableimageheight]{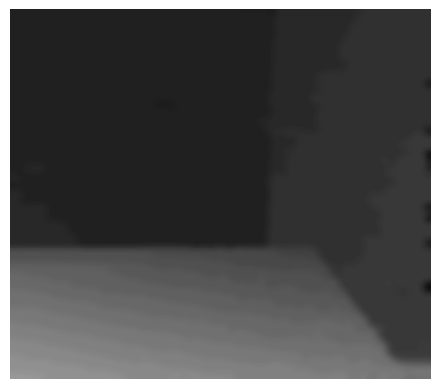}  & \includegraphics[width=\tableimagewidth, height=\tableimageheight]{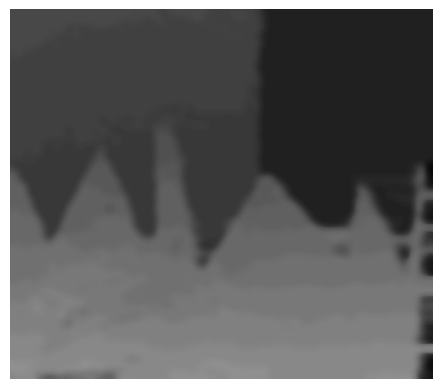} & \includegraphics[width=\tableimagewidth, height=\tableimageheight]{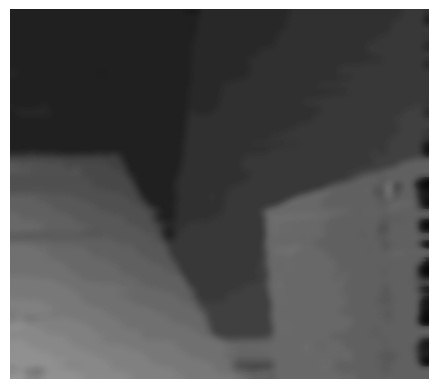}\\
       
    \end{tabular}
    \caption{Visual comparison of our method against \cite{9653310} using \textbf{H}ybrid annealing, and a comparison of our method using \textbf{G}urobi and the method used in \cite{9653310} optimized using the classical \textbf{F}ord Fulkerson algorithm \cite{10.5555/1942094}. For both methods, we can see a marked visual improvement when classical optimizers are used. \textbf{G}urobi provides a near-optimal solution, while \textbf{F}ord Fulkerson provides a global optimum.}
    \label{fig:hybrid_annealing_comp}
\end{figure}

We compare our results visually in \cref{fig:hybrid_annealing_comp} and numerically in \cref{tab:hybrid_annealing_comp}. For added context, we also include the result of using D-Wave's Hybrid optimizer on our method, even though we cannot directly compare to Heidari \etal for the reasons given above. 
In comparison to Heidari \etal's hybrid results, our Gurobi results have significantly fewer streaking artifacts. This can be partially explained by the fact that Heidari \etal's technique only optimizes a single epipolar line at a time. However, because we lack full access to D-Wave's complete hybrid algorithm, we cannot say for certain. When comparing our Gurobi results to Heidari \etal's method optimized with Ford Fulkerson \cite{10.5555/1942094}, we observe that their method has better estimates in the lower left region of Sawtooth than our approach. We suspect this is because Heidari \etal's method has no coarse-to-fine steps, which are leading to estimation errors for Sawtooth (See \cref{sec:coarse_to_fine_ablation} for a full discussion).

\begin{table}[]
    \centering
    \begin{adjustbox}{width=\columnwidth,center}
    \begin{tabular}{|c| c | c || c | c ||| c | c || c | c |}
        \hline
        \multirow{2}{*}{\textbf{Image Pair}} & \multicolumn{2}{ c ||}{\textbf{Ours (\textbf{H})}} & \multicolumn{2}{ c |||}{\textbf{Heidari (\textbf{H})}} & \multicolumn{2}{ c ||}{\textbf{Ours (\textbf{G})}} & \multicolumn{2}{ c |}{\textbf{Heidari (\textbf{F})}} \\ \cline{2-9}
            & RMSE & BPP & RMSE & BPP & RMSE & BPP & RMSE & BPP \\
            \hline
        \textbf{Tsukuba} & 1.79 & 26.82 & 1.8 & 12.8 & 1.53 & 12.93 & 1.58 & 13.02 \\ \hline
        \textbf{Bull}    & 1.66 & 31.57 & 1.3 & 5.4 & 0.58 & 3.46 & 0.56 & 3.57\\ \hline
        \textbf{Sawtooth} & 2.71 & 47.49 & 1.9 & 9.9 & 1.89 & 24.51 & 1.76 & 11.73\\ \hline
        \textbf{Venus}   & 2.16 & 42.43 & 1.4 & 9.8 & 0.96 & 8.16 & 1.17 & 10.25 \\ \hline
        \textbf{Average} &  2.06 & 37.08 & \textbf{1.6} & \textbf{9.48} &  \textbf{1.24} & 12.27 & 1.27 & \textbf{9.40} \\ \hline
        \end{tabular}
    \end{adjustbox}
    \caption{The Root Mean Squared Error (RMSE) and Bad Pixel Percentage (BPP) of our method against \cite{9653310} using \textbf{H}ybrid annealing, and a comparison of our method using \textbf{G}urobi and the method used in \cite{9653310} optimized using the classical \textbf{F}ord Fulkerson algorithm. Under hybrid annealing conditions, the previous method outperforms ours. Under ideal conditions, we can achieve a lower average RMSE, although we have a higher BPP. 
    \label{tab:hybrid_annealing_comp}
    } 
\end{table}

\subsection{Comparison to Traditional Methods}
We also compared our results against non-quantum classical methods assessed in \cite{9653310}, \textit{i.e.,} 
\textbf{B}lock \textbf{M}atching, \cite{BlockMatching}, \textbf{B}elief \textbf{P}ropagation \cite{1206509}, and \textbf{L}ocal-\textbf{E}xpansion \cite{local_expansion_moves}. The numerical results are given in  \cref{tab:classical_method_comparison} and we can see that with an ideal quantum computer, our approach can outperform existing traditional methods. 

\begin{table}[]
    \centering
    \begin{adjustbox}{width=\columnwidth,center}
    \begin{tabular}{|c| c | c || c | c || c | c || c | c |}
        \hline
        \multirow{2}{*}{\textbf{Image Pair}} &  \multicolumn{2}{ c ||}{Ours (\textbf{G})} & \multicolumn{2}{ c ||}{\textbf{BM}} & \multicolumn{2}{ c ||}{\textbf{BP}} &  \multicolumn{2}{ c |}{\textbf{LE}} \\ \cline{2-9}
            & RMSE & BPP & RMSE & BPP & RMSE & BPP & RMSE & BPP \\
            \hline
        \textbf{Tsukuba} & 1.53 & 12.93 & 1.74 & 13 & 1.66 & 9 & 1.01 & 2.9  \\ \hline
        \textbf{Bull}    & 0.58 & 3.46 & 2.76 & 23 & 1.71 & 8 & 0.25 & 0.3   \\ \hline
        \textbf{Sawtooth} & 1.89 & 24.51 & 3.34 & 22 & 1.96 & 10 & 0.81 & 2.8  \\ \hline
        \textbf{Venus}   & 0.96 & 8.16 & 3.27 & 26 & 2.40 & 6 & 0.62 & 2.31 \\ \hline
        \end{tabular}
    \end{adjustbox}
    \caption{The Root Mean Squared (RMSE) and Bad Pixel Percentage (BPP) of our method against the purely classical \textbf{B}lock \textbf{M}atching, \cite{BlockMatching}, \textbf{B}elief \textbf{P}ropagation \cite{1206509}, and \textbf{L}ocal-\textbf{E}xpansion \cite{local_expansion_moves}. We can see that only Local Expansion outperforms our method on RMSE. Except for the Sawtooth pair, we are also comparable to \textbf{BM} and \textbf{BP} on BPP. }
    \label{tab:classical_method_comparison}
\end{table}
\vspace{-10pt}

\subsection{Ablation Studies}
In this section, we perform ablation studies of the elements of our algorithm. We ran our method without a regularization term, with a linear regularizer as in~\cite{9653310}, without bilateral filtering, and without median and bilateral filtering. 

\begin{figure}
    \centering
     \setlength\tabcolsep{1.5pt}
    \begin{tabular}{c c c c c}
                   & Tsukuba & Bull & Sawtooth & Venus \\
        \rotatebox{90}{\phantom{ }Ours (\textbf{G})} & \includegraphics[width=\tableimagewidth, height=\tableimageheight]{sec/images/Best_In_Show_Gurobi/tsukuba.png} & \includegraphics[width=\tableimagewidth, height=\tableimageheight]{sec/images/Best_In_Show_Gurobi/bull.png}  & \includegraphics[width=\tableimagewidth, height=\tableimageheight]{sec/images/Best_In_Show_Gurobi/sawtooth.png} & \includegraphics[width=\tableimagewidth, height=\tableimageheight]{sec/images/Best_In_Show_Gurobi/venus.png}\\
       \rotatebox{90}{\textcolor{white}{---} No \textbf{R}} & \includegraphics[width=\tableimagewidth, height=\tableimageheight]{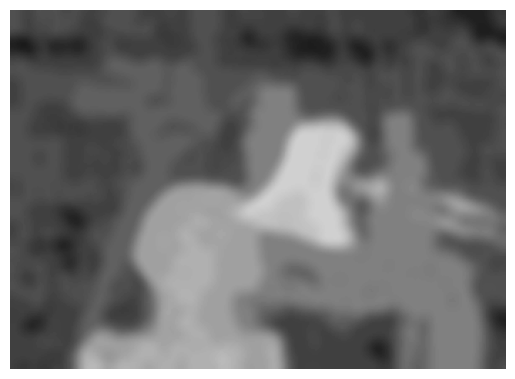} & \includegraphics[width=\tableimagewidth, height=\tableimageheight]{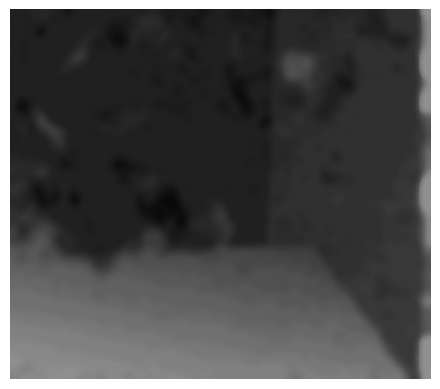}  & \includegraphics[width=\tableimagewidth, height=\tableimageheight]{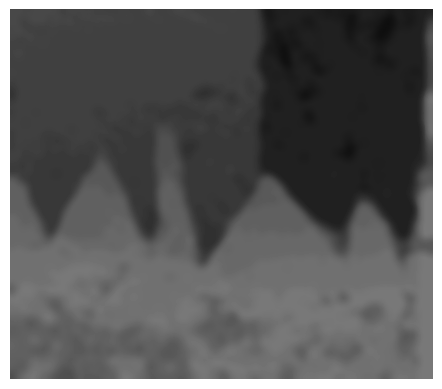} & \includegraphics[width=\tableimagewidth, height=\tableimageheight]{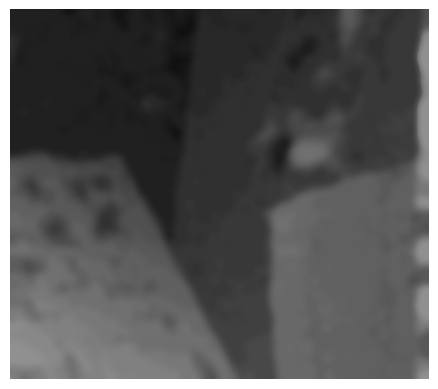}\\
       \rotatebox{90}{\textcolor{white}{-----}\textbf{L R}} & \includegraphics[width=\tableimagewidth, height=\tableimageheight]{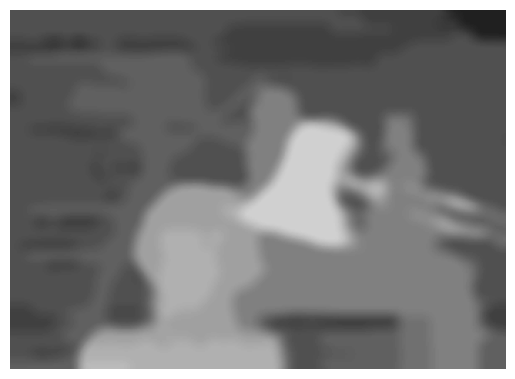} & \includegraphics[width=\tableimagewidth, height=\tableimageheight]{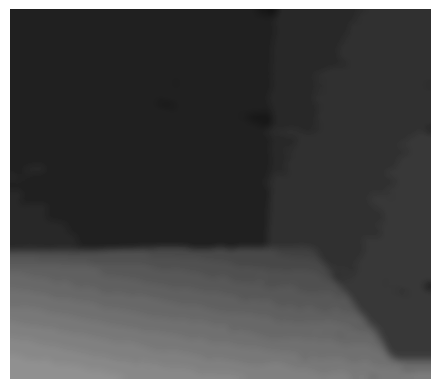}  & \includegraphics[width=\tableimagewidth, height=\tableimageheight]{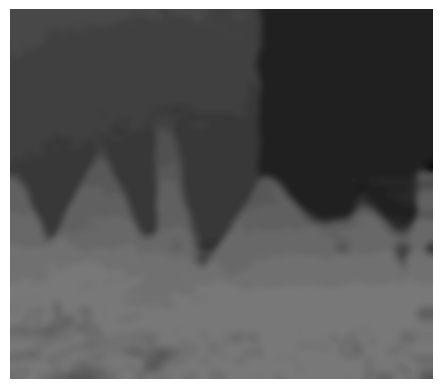} & \includegraphics[width=\tableimagewidth, height=\tableimageheight]{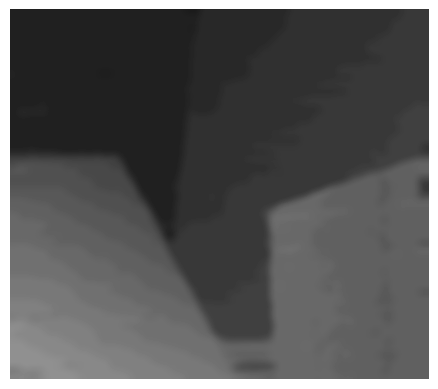}\\
       \rotatebox{90}{\textcolor{white}{-----}No \textbf{B} } & \includegraphics[width=\tableimagewidth, height=\tableimageheight]{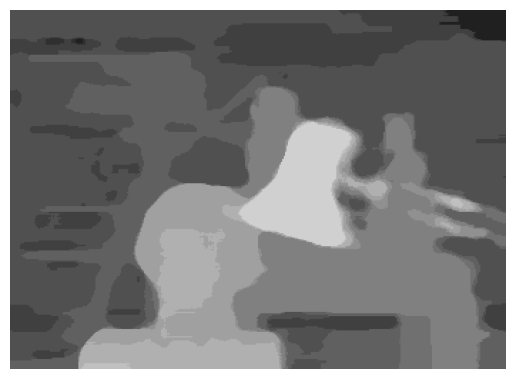} & \includegraphics[width=\tableimagewidth, height=\tableimageheight]{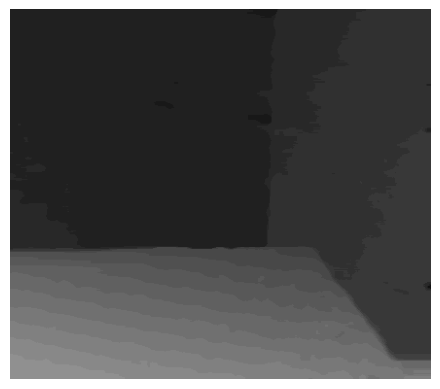}  & \includegraphics[width=\tableimagewidth, height=\tableimageheight]{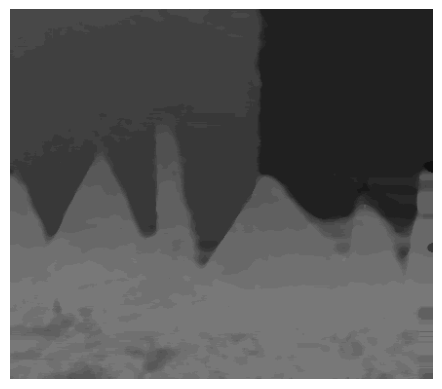} & \includegraphics[width=\tableimagewidth, height=\tableimageheight]{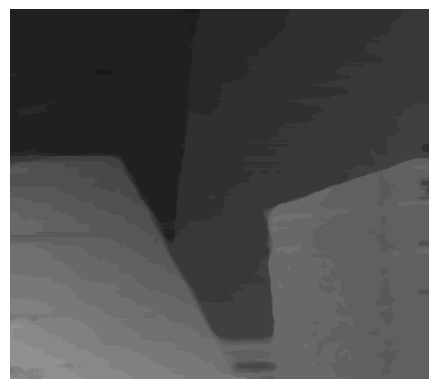}\\
       \rotatebox{90}{No \textbf{M} No \textbf{B}} & \includegraphics[width=\tableimagewidth, height=\tableimageheight]{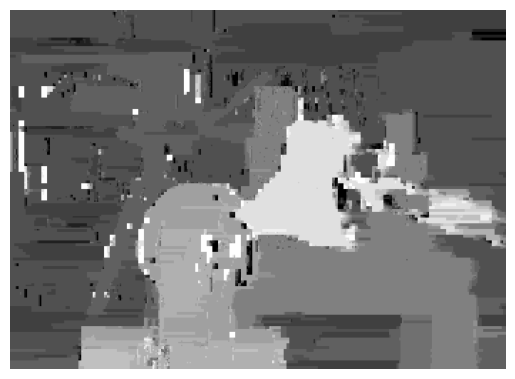} & \includegraphics[width=\tableimagewidth, height=\tableimageheight]{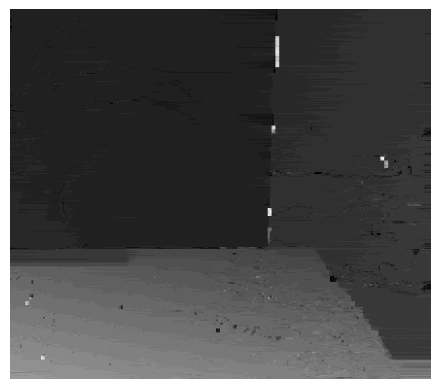}  & \includegraphics[width=\tableimagewidth, height=\tableimageheight]{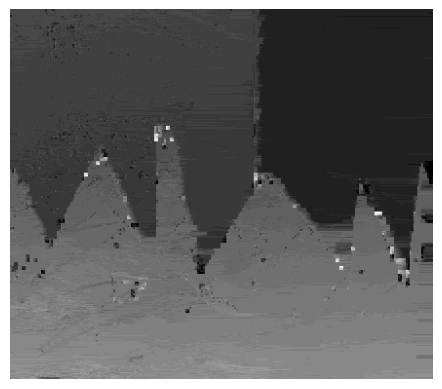} & \includegraphics[width=\tableimagewidth, height=\tableimageheight]{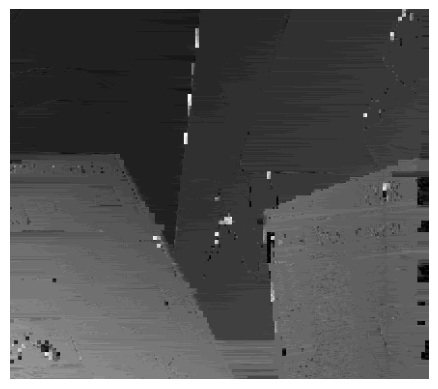}\\
    \end{tabular}
    \caption{
    Ablation study of the components of our algorithm. 
    ``No \textbf{R}egularizer'' sets $E_s{=}0$ from line 12 of \cref{alg:full_quantum_algorithm}, which disables the regularizer and reverts to only using the data term. ``\textbf{L}inear \textbf{R}egularizer'' does not leverage truncation and is not edge-aware. ``No \textbf{B}ilateral filter'' removes the bilateral filtering in line 20 of \cref{alg:full_quantum_algorithm}. ``No \textbf{M}edian and No \textbf{B}ilateral filtering''  row removes lines 18 and 20 from \cref{alg:full_quantum_algorithm}.
    }
    \label{fig:ablation}
\end{figure}

These results are given in~\cref{fig:ablation} and~\cref{tab:ablation} and provide  insights into our method. As expected, when no regularizer is used, the quality drops significantly. Using a linear regularizer instead of our more sophisticated nonlinear regularizer had a small impact on the visual appearance and final outcome. In the case of Bull, it is even slightly lower RMSE than the nonlinear regularizer, although it is still higher on average. We suspect this is due to Bull's disparities being very homogeneous and linear overall, therefore a linear regularizer is advantageous. The bilateral filtering has a blurring effect on the final outcome. Although removing it leads to a small increase RSME, we observe a decrease in BPP. This trade-off occurs because the bilateral filter assists in averaging out the displacements, which generally makes estimates closer in more homogeneous regions, at the cost of a loss of sharpness (and increasing BPP) around the edges. In contrast, removing bilateral and median filtering leads to a much stronger increase in RMSE. This is because median filtering can prevent small inaccuracies made at coarse levels.
\begin{table}[]
    \centering
    \begin{adjustbox}{width=\columnwidth,center}
    \begin{tabular}{|c| c | c || c | c || c | c || c | c || c | c |}
        \hline
        \multirow{2}{*}{\textbf{Image Pair}}  & \multicolumn{2}{ c ||}{Ours (\textbf{G})} & \multicolumn{2}{ c ||}{No \textbf{R}} & \multicolumn{2}{ c ||}{\textbf{L R}} &  \multicolumn{2}{ c ||}{No \textbf{B}} &  \multicolumn{2}{ c |}{No \textbf{M} No \textbf{B}}\\ \cline{2-11}
            & RMSE & BPP & RMSE & BPP & RMSE & BPP & RMSE & BPP & RMSE & BPP \\
            \hline
        \textbf{Tsukuba} & 1.53 & 12.93 & 1.65 & 18.63 & 1.57 & 13.30 & 1.59 & 11.72 & 2.56 & 15.19  \\ \hline
        \textbf{Bull}    & 0.58 & 3.46 & 1.55 & 16.06 & 0.56 & 3.73 & 0.59 & 2.70 & 1.21 & 6.25   \\ \hline
        \textbf{Sawtooth} & 1.89 & 24.51 & 2.30 & 31.54 & 1.95 & 25.20 & 1.95 & 23.41 & 1.99 & 14.13  \\ \hline
        \textbf{Venus}   & 0.96 & 8.16 & 1.64 & 22.22 & 0.96 & 7.50 & 1.03  & 7.23 & 1.89 & 12.34 \\ \hline
        \textbf{Average} & \textbf{1.24} & 12.27 & 1.79 & 22.11 & 1.26 & 12.43 & 1.29 & \textbf{11.27} & 1.91 & 11.98 \\ \hline
        \end{tabular}
    \end{adjustbox}
    \caption{The RMSE and BPP of our method with different elements removed. 
    See the caption of \cref{fig:ablation} for the signification of the abbreviations. 
    }
    \label{tab:ablation}
    \vspace*{-6.2mm}
\end{table}
\section{Conclusion} 

We proposed a new approach for quantum-hybrid stereo matching by formulating it as an MRF MAP inference problem. Thanks to the coarse-to-fine policy, we were able to practically apply our technique to real stereo pairs and achieve higher accuracy than prior quantum-admissible stereo matching methods by $2\%$ to $22.5\%$. As more powerful quantum annealers arise in the future and match or surpass the result quality of Gurobi, our method will directly benefit from these improvements.

The QUBO formulation of a general MRF MAP inference allows for greater flexibility in selecting the smoothness term of our energy functional than prior work. Other problems, such as image segmentation and restoration can be expressed as an MRF MAP estimation and are compatible with our technique. Notably, optical flow can be treated as an extension of stereo matching with a 2D search space that could also be estimated by our framework in the future.
\section*{Acknowledgements} The authors thank Tom Fischer for the draft proofreading. This work was partially funded by the DFG project GRK 2853 ``Neuroexplicit Models of Language, Vision, and Action'' (project number 471607914). 

\clearpage
\twocolumn
{
    \small
    \bibliographystyle{ieeenat_fullname}
    \bibliography{main}
}

\input{sec/X_suppl}
\end{document}